%% file: root.tex
\documentclass[letterpaper, 10 pt, conference]{ieeeconf}  % Comment this line out if you need a4paper

\IEEEoverridecommandlockouts                              % This command is only needed if 
\usepackage[utf8]{inputenc}
\usepackage{graphicx}
\usepackage{amsmath}
\usepackage{amssymb}
\usepackage{algorithm}
\usepackage{algpseudocode}
\usepackage{subfiles}
\usepackage{xcolor}
\usepackage{multirow}
\usepackage{hyperref}
\usepackage[normalem]{ulem}
\graphicspath{{figures/}{../figures/}}

\usepackage{bm}

\title{\LARGE Meta-Learning-Based Robust Adaptive Flight \\ Control Under Uncertain Wind Conditions}
\author{Michael O'Connell, Guanya Shi, Xichen Shi, and Soon-Jo Chung
\thanks{Authors are with California Institute of Technology, USA.
\texttt{\{moc, gshi, xshi, sjchung\}@caltech.edu}.}}

\newcommand{\todo}[1]{{\color{red}}}
\newcommand{\B}[1]{\mathbf{#1}}

\newtheorem{theorem}{Theorem}[section]

\newtheorem{assumption}{Assumption}

\begin{document}

\maketitle
\thispagestyle{plain}
\pagestyle{plain}

\textcolor{blue}{\uline{This article is an early draft and presents preliminary results; the full method and improved results were published in Science Robotics on May 4th, 2022: \href{https://doi.org/10.1126/scirobotics.abm6597}{\texttt{doi.org/10.1126/scirobotics.abm6597}}; arXiv: \href{https://doi.org/10.48550/arXiv.2205.06908}{\texttt{doi.org/10.48550/arXiv.2205.06908}}.}}

\subfile{sections/abstract}

\subfile{sections/intro}

\subfile{sections/problem-statement}

\subfile{sections/meta-learning-kernels}

\subfile{sections/adaptive-control}

\subfile{sections/experiments}

\subfile{sections/discussion}

\section{Acknowledgements}
We thank Yisong Yue, Animashree Anandkumar, Kamyar Azizzadenesheli, Joel Burdick, Mory Gharib, Daniel Pastor Moreno, and Anqi Liu for helpful discussions. The work is funded in part by Caltech's Center for Autonomous Systems and Technologies and Raytheon Company.

% \subsection{Figures and Tables}

% Positioning Figures and Tables: Place figures and tables at the top and bottom of columns. Avoid placing them in the middle of columns. Large figures and tables may span across both columns. Figure captions should be below the figures; table heads should appear above the tables. Insert figures and tables after they are cited in the text. Use the abbreviation ÒFig. 1Ó, even at the beginning of a sentence.

% \begin{table}[h]
% \caption{An Example of a Table}
% \label{table_example}
% \begin{center}
% \begin{tabular}{|c||c|}
% \hline
% One & Two\\
% \hline
% Three & Four\\
% \hline
% \end{tabular}
% \end{center}
% \end{table}

%   \begin{figure}[thpb]
%       \centering
%       \framebox{\parbox{3in}{We suggest that you use a text box to insert a graphic (which is ideally a 300 dpi TIFF or EPS file, with all fonts embedded) because, in an document, this method is somewhat more stable than directly inserting a picture.
% }}
%       %\includegraphics[scale=1.0]{figurefile}
%       \caption{Inductance of oscillation winding on amorphous
%       magnetic core versus DC bias magnetic field}
%       \label{figurelabel}
%   \end{figure}

\addtolength{\textheight}{-12cm}   % This command serves to balance the column lengths



% \section*{ACKNOWLEDGMENT}

% The authors thank 

\bibliographystyle{IEEEtran}
\bibliography{IEEEabrv,ref}

\end{document}

%% file: sections/abstract.tex
\begin{abstract}
Realtime model learning proves challenging for complex dynamical systems, such as drones flying in variable wind conditions. Machine learning technique such as deep neural networks have high representation power but is often too slow to update onboard. On the other hand, adaptive control relies on simple linear parameter models can update as fast as the feedback control loop. We propose an online composite adaptation method that treats outputs from a deep neural network as a set of basis functions capable of representing different wind conditions. To help with training, meta-learning techniques are used to optimize the network output useful for adaptation. We validate our approach by flying a drone in an open air wind tunnel under varying wind conditions and along challenging trajectories. We compare the result with other adaptive controller with different basis function sets and show improvement over tracking and prediction errors.
\end{abstract}

%% file: sections/intro.tex
\section{Introduction}
For a given dynamical system, complexity and uncertainty can arise either from its inherent property or the changing environment. Thus model accuracy is often key in designing a high-performance and robust control system. If the model structure is known, conventional system identification techniques can be used to resolve the parameters of the model. When the system becomes too complex to model analytically, modern machine learning research conscripts data-driven and neural network approaches that often result in bleeding-edge performance given enough samples, proper tuning, and adequate time for training. However, the harsh requirement on a learning-based control system calls for both representation power and fast in execution simultaneously. Thus it is natural to seek wisdom from the classic field of adaptive control, where successes have been seen using simple linear-in-parameter models with provably robust control designs~\cite{slotine1991applied,farrell2006adaptive}. On the other hand, the field of machine learning has made its own progress toward fast online paradigm, with the rising interest in few-shot learning~\cite{snell2017prototypical}, continual learning~\cite{kirkpatrick2017overcoming,zenke2017continual}, and meta learning~\cite{santoro2016meta,finn2017model}.

A particular interesting scenario for a system in a changing environment is a multi-rotor flying in varying wind conditions. Classic multi-rotor control does not consider the aerodynamic forces such as drag or ground effects~\cite{mellinger2011minimum}. The thruster direction is controlled to follow the desired acceleration along a trajectory. To account for aerodynamic forces in practice, an integral term is often added to the velocity controller~\cite{meier2012pixhawk}. Recently, \cite{tal2018accurate} uses incremental nonlinear dynamic inversion (INDI) to estimate external force through filtered accelerometer measurement, and then apply direct force cancellation in the controller. \cite{faessler2017differential} assumed a diagonal rotor drag model and proved differential flatness of the system for cancellation, and \cite{shi2018nonlinear} used a nonlinear aerodynamic model for force prediction. When a linear-in-parameter (LIP) model is available, adaptive control theories can be applied for controller synthesis. This does not limit the model to only physics-based parameterizations, and a neural network basis can be used~\cite{nakanishi2002locally,farrell2006adaptive}. It has been applied to multi-rotor for wind disturbance rejection in~\cite{bisheban2019geometric}.
\begin{figure}[t]
\label{fig:wind-tunnel}
\includegraphics[width=\linewidth]{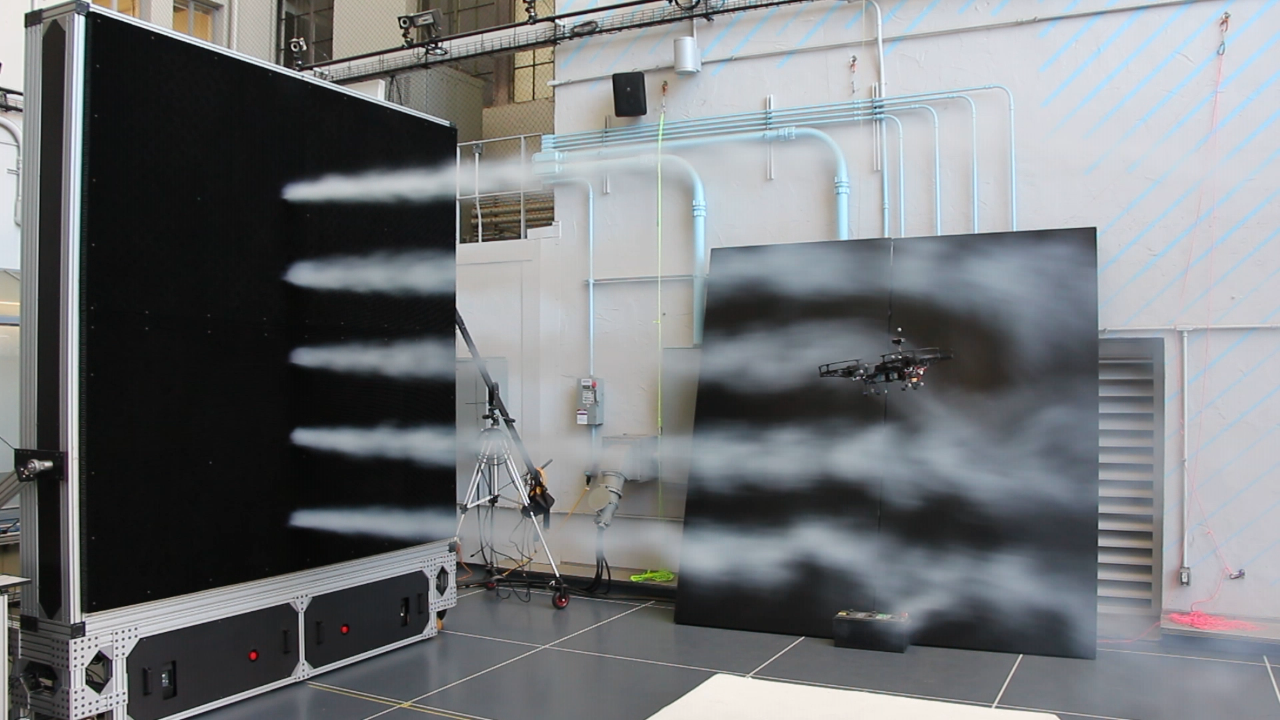}
\caption{Intel Aero drone flying in front of Caltech CAST Fan Array.}
\label{fig:intro}
\end{figure}

When adapting to complex system dynamics or fast changing environment, one would expect the network approximator to have enough representation power, which makes a deep neural network (DNN) an desirable candidate. However, there are several issues associated with using a deep network for adaptive control purpose. First, training a DNN often requires back propagation, easily leading to a computation bottleneck for realtime control on small drones. Second, continual online training may incur catastrophic inference where previously learned knowledge is forgotten unintentionally. Third, a vanilla network for a regression problem often does not have guarantees on desirable properties for control design, such as output boundedness and Lipschitz continuity. Fortunately, advances have been made in circumventing these issues. Training a deep network by updating the last layer's weight more frequently than the rest of the network is proven to work for approximating $Q$-function in reinforcement learning~\cite{levine2017shallow,azizzadenesheli2018efficient}. This enables the possibility of fast adaptation without incurring high computation burden. Spectral normalization on all the network weights can constrain the Lipschitz constant~\cite{miyato2018spectral}. We used this technique in our prior work~\cite{shi2018neural} to derive stable DNN-based controller for multirotor landing.

We thus propose the following method for an online composite adaptive control based on DNN. We approximate the unknown part of our dynamics model with a DNN trained offline with previously collected data. When deployed online, only the last layer weights are updated in a fashion similar to composite adaptive control~\cite{slotine1989composite}. The training process employs model-agnostic meta-learning (MAML) technique from~\cite{finn2017model} to facilitate hidden layer outputs becoming good basis functions for online adaptation. All the network weights are spectrally normalized during training as well as online adaptation, to constrain approximator Lipschitz constant, which was proven to be a necessary condition for stable control design~\cite{shi2018neural}.

%% file: sections/problem-statement.tex
\section{{Problem Statement}}
\subsection{Mixed Model for Robot Dynamics}
Consider the general robot dynamics model:
\begin{equation}
    \label{equ:open-loop-dynamics}
    H(q)\ddot q + C(q,\dot q) \dot q + g(q) + f(q,\dot q; c) = \tau
\end{equation}
where $q, \dot q, \ddot q \in \mathbb{R}^n$ are the $n$ dimensional position, velocity, and acceleration vectors, $H(q)$ is the symmetric, positive definite inertia matrix, $C(q,\dot q)\dot q$ is the centripetal and Coriolis torque vector, $g(q)$ is the gravitational torque vector, $f(q,\dot q; c)$ incorporates unmodeled dynamics, and $c=c(t)$ is hidden state used to represent changing envirnoment. 

We approximate the unmodeled dynamics term by a linear combination of a set of $m$ neural network kernels. We consdier two formulations here. First, we approximate $f(q,\dot q; c)$ by linearly combining $m$ outputs from $m$ separately trained neural networks $\vec{\varphi}_i : \mathbb{R}^n \times \mathbb{R}^n \rightarrow \mathbb{R}^n$ parameterized by $\theta_i$:
\begin{equation}
    \label{equ:meta-1-kernels}
    f(q, \dot q; c) \approx \sum_{i=1}^m  a_i(c)\vec{\varphi}_i(q, \dot q; \theta_i) = \varphi(q, \dot q; \Theta) a(c)
\end{equation}
where $a(c) = [a_i(c)] \in \mathbb{R}^n$ and the kernels are stacked such that $\varphi(q, \dot q; \Theta) = [\vec{\varphi}(q,\dot{q}, \theta_i)_i]$ and $\Theta = [\theta_i]$. 

Second, we consider the alternative formulation where $f(q, \dot q; c)$ is approximated with a single neural network, where $a$ represents the weights of its last layer, and $\{\varphi_i\}$ represent the hidden states before the last layer. This can be explicitly written as
\begin{align}
    \label{equ:meta-2-kernels}
    f(q, \dot q; c) &\approx \sum_{i=1}^m \sum_{j=1}^n  a_{i,j}(c)\varphi_i(q, \dot q; \theta_i) \hat{\mathrm{e}}_j = \varphi(q, \dot q; \Theta) a(c) \nonumber \\ 
    \varphi(q, \dot q; \Theta) &\triangleq \begin{bmatrix} [\varphi_i]^T & 0 & 0 \\ 0 & [\varphi_i]^T & 0 \\ 0 & 0 & [\varphi_i]^T\end{bmatrix}
\end{align}\todo{generalize for n!=3}
where $\hat{\mathrm{e}}_j$ represent the standard basis vectors.

In both cases, maximum representation error, $\epsilon$, is
\begin{equation}
    \label{equ:epsilon-defintion}
    \epsilon = \min_{a \in \mathbb{R}^m}\max_{(q, \dot q, c) \in \Xi}\|\varphi(q, \dot q; \Theta)a - f(q, \dot q; c)\|
\end{equation}
where $\Xi$ is the compact domain of interest. Note, the boundedness of $\epsilon$ is apparent under the assumption of bounded Lipschitz constant of $f(q, \dot q, c)$ and bounded training error. %, since the distance between the the test trajectory and the training data is bounded and the kernel functions have bounded Lipschitz constant, which is ensured during the offline training.
Given $\Theta$, the goal then is to design a control law, $\tau(q, \dot q, q_d, \dot q_d)$, that drives $(q, \dot q) \to (q_d, \dot q_d)$, subject to dynamics in (\ref{equ:open-loop-dynamics}).
%%%%%%%%%%%%%%%%%%%%%%%%%%%%%%%%%%%%%%%%%%%%%%%%%%%%%%%%%%%%%%%%%%%%%%%%%%%
\subsection{Quadrotor Position Control Subject to Uncertain Wind Conditions}
Now we specialize the problem to quadrotors. Consider states given by global position, $p \in \mathbb{R}^3$, velocity $\B{v} \in \mathbb{R}^3$, attitude rotation matrix $R \in \mathrm{SO}(3)$, and body angular velocity $\omega \in \mathbb{R}^3$. Then dynamics of a quadrotor is
\begin{subequations}
\begin{align}
\dot{p} &= v, &  
m\dot{v} &=mg+Rf_u + f_a,\label{eq:pos-dynamics} \\ 
\dot{R}&=RS(\omega), & 
J\dot{\omega} &= J \omega \times \omega  + \tau_u + \tau_a,
\label{eq:att-dynamics}
\end{align}
\label{eq:dynamics}
\end{subequations}
where $m$ is the mass and $J$ is the inertia matrix of the system, $S(\cdot)$ is the skew-symmetric mapping, $g = [0, 0, -g]^\top$ is the gravity vector, $f_u = [0, 0, T]^T$ and $\tau_u = [\tau_x, \tau_y, \tau_z]^T$ are the total thrust and body torques from four rotors predicted by a nominal model, and $f_a = [0, 0, T]^T$ and $\tau_a = [\tau_x, \tau_y, \tau_z]^T$ are forces and torques resulting from unmodelled aerodynamic effects due to varying wind conditions.  

% We use $\eta = [T, \tau_x, \tau_y, \tau_z]^\top$ to denote the output wrench. Typical quadrotor control input uses squared motor speeds $u = [n_1^2,n_2^2,n_3^2,n_4^2]^T$, and is linearly related to the output wrench $\eta = B_0 u$, with
% \begin{equation}
% B_0=
% \left[\begin{smallmatrix}
% c_T & c_T & c_T & c_T \\
% 0 & c_T l_{\mathrm{arm}} & 0 & - c_T l_{\mathrm{arm}}\\
% -c_T l_{\mathrm{arm}} & 0 & c_T l_{\mathrm{arm}} & 0 \\
% -c_Q & c_Q & -c_Q & c_Q
% \end{smallmatrix}\right],
% \label{eq:bmatrix}
% \end{equation}
% where $c_T$ and $c_Q$ are rotor force and torque coefficients, and $l_{\mathrm{arm}}$ denotes the length of rotor arm. 

For drones in strong wind conditions, the primary disturbance is unmodelled aerodynamic forces $f_a=[f_{a,x}, f_{a,y}, f_{a,z}]^\top$.  Thus, considering only position dynamics, we cast (\ref{eq:pos-dynamics}) into the form of (\ref{equ:open-loop-dynamics}), by taking $H(q) = m \mathrm{I}$, where $\mathrm{I}$ is the identity matrix, $C(q, \dot q) \equiv 0$, $g = m \B{g}$, $f(q, \dot q; c) = \B{f}_a$, and $\tau = R\B{f}_u$.  Note that the quadrotor attitude dynamics is just a special case of~\eqref{equ:open-loop-dynamics}.

\subsection{Meta-Learning and Adaptation Goal}
Suppose we have pre-collected meta-training data $D_{\text{meta}}=\{D_1,D_2,\cdots,D_T\}$, with $T$ sub datasets. In each sub dataset, 
% we have 
% \begin{equation}
% D_i = \{([q_k,\dot{q}_k],\hat{f}(q_k,\dot{q}_k,c_i))\}_{k=1}^{L_i}.
% \end{equation}
% that is, 
$D_i$, we have $L_i$ state and force measurement pairs, $([q_k,\dot{q}_k],\hat{f}(q_k,\dot{q}_k,c_i))$, generated from some fixed but unknown wind condition, represented by $c_i$. The goal of meta-learning is to generate a set of parameters, $\Theta = \{\theta_i\}_{i=1}^m$, such that a linear combination  of the neural net kernels, $\{\varphi_i\}_{i=1}^m$, can represent any wind condition with small error.  %This process of learning $\Theta$ is called meta-learning.  In this case, $\Theta$ and $\epsilon$ will become the prior for the wind conditions that the drone might encounter in the control problem.

Consequently, the adaptive controller aims to stabilize the system to a desired trajectory given the prior information of the dynamic model (\ref{equ:open-loop-dynamics}) and the learned kernels, (\ref{equ:meta-1-kernels}) or (\ref{equ:meta-2-kernels}). If exponential convergence is guaranteed, the closed-loop system to aerodynamic effects not captured by the prior from meta learning, encapsulated in $\epsilon$.  %By incorporating the learned kernel functions and adaptively updating the linear combination of those kernels, we will have systematically improve flight performance over a controller that just uses the modelled flight dynamics.  

% How do we define good prior/kernels? This question would be clear in the adaptation phase. Suppose we meta-learned $\Theta^*$ from $D_{\text{meta}}$. We will have few data points $\{([q_k,\dot{q}_k],\hat{f}(q_k,\dot{q}_k,c_{\text{new}}))\}_{k=1}^K$ online, generated by some new wind condition $c_{\text{new}}$. With these $K$ training data points, we will online learn $\hat{a}(c_{\text{new}})$. The definition of good prior/kernels is that
% \begin{equation}
% \|\varphi(q,\dot{q},\Theta^*)\hat{a}(c_{\text{new}})-f(q,\dot{q},c_{\text{new}})\|
% \end{equation}
% is as small as possible.

% In summary, the goal of meta-learning is to learn $\Theta^*$, such that we could do fast adaptation with these learned kernels, $\varphi(q,\dot{q},\Theta^*)$. Note that once meta-learning is finished, we would fix $\Theta^*$, and only $a$ will be online learned.

%% file: sections/meta-learning-kernels.tex
\section{{Meta-Learning Neural Net Kernels}}
\label{sec:learning}

\todo{add spectral normalization}
\todo{emphasize spectral normalization}
Recall that the meta learning goal is to learn a set of kernel functions, $\{\varphi(q, \dot q)\}$, such that for any wind condition, $c$, there exists a suitable $a$ such that $\varphi(q, \dot q, \Theta)$ is a good approximation of $f(q, \dot q, c)$.  We can formulate this problem as the minimization problem
\begin{equation}
    \label{equ:learning-cost-function}
    \Theta^* = \min_{a_i, \Theta} \sum_i J_i(a_i, \Theta)
\end{equation}
%with
\begin{align}
    \text{with   }& J_i(a_i, \Theta, (q_i(\cdot), \dot q_i(\cdot))) \\&= \int_0^{t_f} \| f(q_i(r), \dot q_i(r), c_i) - \varphi(q_i(r), \dot q_i(r), \Theta) a_i \| ^ 2 dr \nonumber \\
    &\approx J(a_i, \Theta, D_i) = \Delta t \cdot \sum_{(q, \dot q, f) \in D_i} \| f - \varphi(q, \dot q), \Theta) a_i \| ^ 2\nonumber
\end{align}
where the training data is divided into subsets, $D_i$, each corresponding to a fixed wind conditions, 

Note that (\ref{equ:learning-cost-function}) can be equivalently written as 
\begin{equation}
    \label{equ:learning-cost-function-split}
    \min_{\Theta} \sum_i \min_{a_i} J_i(a_i, \Theta)
\end{equation}
The inner problem, $\sum_i \min_{a_i} J_i(a_i, \Theta)$, is simply a linear least squares problem and can be solved exactly for a fixed value of $\Theta$.  Since there are many more training examples, given as discrete measurements of $(q, \dot q, f(q, \dot q, c)$, than parameters, $a$, the least squares solution is over determined and we can approximate it well with the least squares solution on a small, randomly sampled subset of the training examples, $D_i^a$.  The remaining examples $D_i^\Theta$, such that $D_i^a \cup D_i^\Theta = D_i$ and $D_i^a \cap D_i^\Theta = \emptyset$.  

Write the least squares solution for $a$ as
\begin{equation}
    a=a_{LS}(\Theta,D_i^{\text{a}}).
\end{equation}
Note that this solution can be explicitly written as the solution to the following equation.
\begin{equation}
\underbrace{\begin{bmatrix}\varphi(q^{(1)}, \dot q^{(1)},\theta)\\
\varphi(q^{(2)}, \dot q^{(2)},\theta) \\
\vdots \\
\varphi(q^{(K)}, \dot q^{(K)}, \theta)
\end{bmatrix}}_{\Phi\in\mathbb{R}^{K\times m}} a = \underbrace{\begin{bmatrix}
f^{(1)} \\ f^{(2)} \\ \vdots \\ f^{(K)}
\end{bmatrix}}_{F}.
\end{equation} 
where $K$ is the size of $D_i^a$.
Therefore the least-square solution will be
\begin{equation}
a_{LS}(\Theta, D_i^a) = LS(\Theta,D_i^{\text{a}}) = (\Phi^T\Phi)^{-1}\Phi^T F
\end{equation}

Now with $a$ as a function of $\Theta$, we can solve the outer problem in (\ref{equ:learning-cost-function-split}) using stochastic gradient descent on $\Theta$.  This gives the following iterative algorithm for solving for $\Theta$.

% \begin{alg}
%     \label{alg:meta-learning}
%     Randomly initialize $\Theta_0$.  For each dataset $D_i$, randomly split into two subsets $(D_i^a, D_i^\Theta)$.  Update $\Theta$ as
%     \begin{equation}
%         \Theta \gets \Theta-\beta\cdot\nabla_{\Theta} J_i \left( a_{LS} ( \Theta, D_i^a
%         ), \Theta, D_i^{\Theta}\right)
%     \end{equation}
%     Repeat partitioning and $\Theta$ update until $\Theta$ converges.
% \end{alg}

\begin{algorithm}
    \label{alg:meta-learning}
    \caption{Meta-Learning Algorithm}% \label{alg:meta-learning}
    \begin{algorithmic}[1]
    \Procedure{Meta-Learning}{$\{D_i\}$}\Comment{$\Theta^*$}
        \State \textbf{initialize} $\Theta_0$ randomly, $k = 0$ 
        \Repeat
            \For{$ i \in \{1, ... T\}$}
                \State ($D_i^a, D_i^\Theta) \gets \text{random split } D_i $
                \State $\Theta_{k+1} =\Theta_{k}$ 
                \Statex $\quad\quad\quad\quad\quad - \beta \cdot \nabla_{\Theta_{k}} J_i \left( a_{LS} ( \Theta_{k}, D_i^a), \Theta_{k} , D_i^\Theta \right)$
            \EndFor
            \State $k \gets k+1 $
        \Until{converged}
    \EndProcedure
    \State \textbf{return} $\Theta_k$
    \end{algorithmic}
\end{algorithm}
% \begin{theorem}
%     Approximate convergence of Algorithm \ref{alg:meta-learning}
% \end{theorem}
% \begin{proof}
    Note, if $D_i^a = D_i^\Theta = D_i$ then asymptotic convergence is guaranteed since solving the least squares problem is monotonically decreasing during each iteration, the batch update law is monotonically decreasing for small enough $\beta$, and the 2-norm, and hence the cost, is lower bounded by 0.  
    % \end{proof}

%% file: sections/adaptive-control.tex
\section{{Robust Composite Adaptation}}
\label{sec:control}

Recall the control design objective is to design a control system that leverages the kernels, $\varphi(q, \dot q; \Theta)$, to stabilize the system defined in (\ref{equ:open-loop-dynamics}), to some desired trajectory $(q_d, \dot q_d)$.  Treating $\Theta$ as fixed, we will not notate dependence on $\Theta$ in this section.  The control system will have two parts: a control law, $\tau(q, \dot q, q_d, \dot q_d, \hat a)$, and an update law, $\hat a(q, \dot q, q_d, \dot q_d, \tau)$.

In the process of designing the control system, we make a few key assumptions.
\begin{assumption}
\label{ass:bounded-traj}
The desired trajectory and its first and second derivative, $\{q_d(t), \dot q_d(t), \ddot q_d(t)\}$, are bounded.
\end{assumption}
\begin{assumption}
\label{ass:representation-error}
The flown flight trajectory, $(q(t), \dot q(t))$, and the current wind conditions, $c$, are a subset of $\Xi$.  Thus, the optimal parameters for the flown flight trajectory and current wind conditions, given by $a = a_{LS}(\Theta, (q(t), \dot q(t), f(q(t), \dot q(t), c))$, with pointwise representation error, $d(q, \dot q) = \|\varphi(q, \dot q; \Theta)a - f(q, \dot q; c)\|$, have maximum representation error along the flown flight trajectory, $d$, less than the maximum global representation error, $\epsilon$.  That is,
\begin{equation}
    d(q, \dot q) \triangleq \|\varphi(q, \dot q; \Theta)a - f(q, \dot q; c)\| \leq d 
\end{equation}
\begin{equation} 
    \label{equ:d-definition}
    d \triangleq \min_{a\in\mathbb{R}^m}\max_{t} \|\varphi(q(t), \dot q(t)) a - f(q(t), \dot q(t), c)\| \leq \epsilon
\end{equation}

\end{assumption}

Note that for time varying optimal parameters, $a = a(t)$, we can follow the same formulation but have an additional disturbance term proportional to $\dot a$.  %Thus, for slowly varying optimal parameters where the disturbance $\dot a$ is small, the following method will suffice.  

\subsection{Nonlinear Control Law}

In formulating our control problem, we first define the composite velocity tracking error term, $s$, and the reference velocity, $\dot q_r$, such that $
    s = \dot q - \dot q_r = \dot{\tilde{q}} + \Lambda \tilde{q}
$, where $\tilde{q} = q - q_d$ is the position tracking error, and $\Lambda$ is a control gain and positive definite.  Then given parameter estimate $\hat a$, we define the following control law
\begin{equation}
    \label{equ:control-law}
    \tau = H(q) \ddot{q_r} + C(q, \dot q) \dot q_r + g(q) + \varphi(q, \dot q) \hat a - K s
\end{equation}
where $K$ is another positive definite control gain.  Combining (\ref{equ:open-loop-dynamics}) and (\ref{equ:control-law}) leads to the closed-loop dynamics of 
\begin{equation}
    \label{equ:composite-error-dynamics}
    H(q) \dot s + (C(q, \dot q) + K) s - \varphi(q, \dot q) \tilde{a} = d(q, \dot q)
\end{equation}

\subsection{Composite Adaptation Law}
We will define an adaptation law that combines a tracking error update term, a prediction error update term, and a regularization term.  This formulation follows \cite{slotine1989composite} with the inclusion of regularization.  

First, we define the prediction error as
\begin{equation}
    \label{equ:prediction-error}
    e(q, \dot q) \triangleq \varphi(q,\dot q) \hat a - f(q, \dot q, c) = \varphi(q, \dot q) \tilde{a} + d(q, \dot q)
\end{equation}
Next, we filter the right hand side of (\ref{equ:prediction-error}) with a stable first-order filter with step response $w(r)$ to define the filtered prediction error.  
\begin{equation}
    e_1(\hat a, t) \triangleq W(t) \hat a - y_1(t) = W(t) \tilde a + d_1(t)
\end{equation}
with filtered measurement, $y_1 = y_1(t) = \int_0^t w(t-r) y(r) dr$, filtered kernel function, $W = W(t) = \int_0^t w(t-r) \varphi(r) dr$, and filtered disturbance, $d_1 = d_1(t) = \int_0^t w(t-r) d(r) dr$.\todo{the same symbols repeated (W=W(t)). Do you need another symbol for a filtered version?}

Now consider the following cost function.
\begin{equation}
    \label{equ:J2}
    J_2(\hat a) = \int_0^t \mathrm{e}^{-\lambda (t-r)} \|W(r) \hat a - y_1(r)\|^2 dr + \gamma \|\hat a\|^2
\end{equation}
Note this is is closely related to the cost function defined in (\ref{equ:learning-cost-function}), with three modifications.  First, the inclusion of an exponential forgetting factor will lead to exponential convergence of $\sqrt{\Bar{P}^{-1}} \tilde{a}$, where $\Bar{P}$ will be defined in (\ref{equ:pbar-definition}).  Second, the regularization term, $\gamma \|\hat a\|^2$ will guarantee invertibility of $\Bar{P}$, even without the persistence of excitation condition usually required to guarantee parameter convergence.  However, note that lack of persistence of excitation could lead to poorly conditioned $\Bar{P}^{-1}$.  Regularization also introduces an additional disturbance term proportional to $a$, as seen later in (\ref{equ:stacked-dynamics}).  Third, this cost function uses the filtered prediction error instead of the unfiltered prediction error to smooth the update law and to allow use to remove $\ddot q$ from the update law via integration by parts on $y_1$.  

Note that $J_2$ is quadratic and convex in $\hat a$, leading to a simple closed form solution for $\hat a$.  However, this requires evaluating an integral over the entire trajectory at every time step, so differentiating this closed form solution for $\hat a$ gives the following prediction error with regularization regularization update law.
\begin{align}
    \label{equ:pred-error-adaptation}
    \dot{\hat a} &= - \Bar{P} ( W^T e_1 + \lambda \gamma \hat a ) \\
    \label{equ:pbar-inv-derivative}
    \dot{\Bar{P}} &= \left( \lambda - \lambda \gamma \Bar{P} - \Bar{P} W^T W\right) \Bar{P}
\end{align}
where
\begin{equation}
    \label{equ:pbar-definition}
    \Bar{P} \triangleq \left( \int_0^t \mathrm{e}^{-\lambda (t-r)} W^T W dr + \gamma \right)^{-1}
\end{equation}

Now, we define the composite adaptation law with regularization, which incorporates an additional tracking error based term proportional to $s$ into (\ref{equ:pred-error-adaptation}).  Later, we will see that this tracking error term exactly cancels the $\tilde{a}$ term in  (\ref{equ:composite-error-dynamics}).
\begin{equation}
    \label{equ:adaptation-law}
    \dot{\hat a} = - \Bar{P}( %\Gamma \Bar{P}
    \varphi^T s + W^T e_1 + \lambda \gamma \hat a)
\end{equation}
%where $\Gamma$ is an adaptation gains.d

% \begin{theorem}{Algorithm (Meta-Learning Optimal Kernels}  

% \end{theorem}
% \begin{proof}
% blah
% \end{proof}
\begin{theorem}
    Under Assumptions~\ref{ass:bounded-traj} and~\ref{ass:representation-error} and using the control law defined in (\ref{equ:control-law}), the composite tracking error and parameter estimation error evolving according to the dynamics in (\ref{equ:composite-error-dynamics}) and adaptation law in (\ref{equ:pbar-inv-derivative}-\ref{equ:adaptation-law}) exponentially converge to the error ball
    \begin{equation}
        \label{equ:exponential-bound}
        \left\| \begin{bmatrix} s\\ \tilde{a}\end{bmatrix} \right\| 
        \leq 
        \frac{\sup_t[\kappa(H(q(t))), \kappa(\Bar{P}^{-1})]}
        {\min[k, \frac{\lambda}{2}(\lambda_\mathrm{min}(\Bar{P}^{-1}) + \gamma)] } \sup \left\| 
        \begin{bmatrix}
            d \\ %-\Gamma^{-1} 
            W^T d_1 \!+\! \lambda \gamma a
        \end{bmatrix}\right\| 
    \end{equation} 
where $k = \lambda_{\min}K$ and $\kappa(\cdot)$ is the condition number.
    \end{theorem}
\begin{proof}
    Rearranging the composite tracking error dynamics and the parameter estimate dynamics, defined in (\ref{equ:composite-error-dynamics}) and (\ref{equ:adaptation-law}), and using the derivative of $\Bar{P}^{-1}$ given in (\ref{equ:pbar-inv-derivative}), we get the combined closed loop dynamics
    \begin{multline}
        \label{equ:stacked-dynamics}
        \begin{bmatrix}
            H(q) & 0 \\
            0 & \Bar{P}^{-1} %\Gamma^{-1}
        \end{bmatrix}
        \begin{bmatrix}
            \dot s \\ \dot {\tilde{a}}
        \end{bmatrix} 
        + 
        \begin{bmatrix}
            C(q, \dot q) + K & -\varphi(q, \dot q) \\
            \varphi(q, \dot q)^T & W^T W + \lambda \gamma I \\ %\Gamma^{-1} \\
        \end{bmatrix}
        \begin{bmatrix}
            s \\ \tilde{a}
        \end{bmatrix} \\
        =
        \begin{bmatrix}
            d(q, \dot q) \\ - % \Gamma^{-1}
            (W^T d_1 + \lambda \gamma a)
        \end{bmatrix}
    \end{multline}
    
    Consider the Lyapunov-like function $V = y^T M y$, with $y = \begin{bmatrix} s^T & \tilde{a}^T\end{bmatrix}^T$ and metric function, $M$, given by 
    \begin{equation}
        M = \begin{bmatrix}
            H(q) & 0 \\
            0 & \Bar{P}^{-1}% \Gamma^{-1}
        \end{bmatrix}
    \end{equation}
    Using the closed loop dynamics given in (\ref{equ:stacked-dynamics}) and the skew symmetric property of $\dot M-2C$, we get the following inequality relationship the derivative of $V$.
    \begin{align}
        \frac{d}{dt}(y^T  M y) 
        &= 2 y^T M \dot y + y^T \dot M y \\
        % &= y^T \left( -2
        %     \begin{bmatrix}
        %         C(q, \dot q) + K & -\varphi(q, \dot q) \\
        %         \varphi(q, \dot q)^T & W^T W + \lambda \gamma I \\
        %     \end{bmatrix}  y \right. \\ & \quad \left.
        %     + 
        %     \begin{bmatrix}
        %         d(q, \dot q) \\ -(W^T D + \lambda \gamma a)
        %     \end{bmatrix}
        % \right)
        %     + 
        %     y^T
        %     \begin{bmatrix}
        %         \dot H & 0 \\ 0 & -\lambda \Bar{P}^{-1} + \lambda \gamma I + W^T W
        %     \end{bmatrix}
        %     y \\
        &= -y^T 
            \begin{bmatrix}
                2 K & 0 \\
                0 & \lambda \Bar{P}^{-1} + \lambda \gamma I + W^T W
            \end{bmatrix}
            y 
            \nonumber\\&\quad 
            + y^T
            \begin{bmatrix}
                d \\ - %\Gamma^{-1}
                (W^T d_1 + \lambda \gamma a)
            \end{bmatrix}\\
        % &\leq 
        %     -\frac
        %         {\min[2 k, \lambda_\mathrm{min}(\lambda \Bar{P}^{-1} + \lambda \gamma I + W^T W )]}
        %         {\lambda_\mathrm{max}(M)} y^T M y \\ &\quad 
        %     + \sqrt{\frac{y^T M y}{\lambda_\mathrm{min}(M)}}
        %     \left\| 
        %     \begin{bmatrix}
        %         d \\ W^T D + \lambda \gamma a
        %     \end{bmatrix}\right\| \\
        % &\leq 
        % -2 \frac
        %     {\min[k, \frac{1}{2}\lambda_\mathrm{min}(\lambda \Bar{P}^{-1}) + \frac{1}{2} \lambda \gamma]}
        %     {\lambda_\mathrm{max}(M)} y^T M y \nonumber\\ &\quad 
        % + \sqrt{\frac{y^T M y}{\lambda_\mathrm{min}(M)}}
        % \left\| 
        % \begin{bmatrix}
        %     d \\ W^T D + \lambda \gamma a
        % \end{bmatrix}\right\| \\
        % &=
        % -2 \lambda_{con} y^T M y + \Bar{d} \sqrt{y^T M y}\\
        % &= 
        &\leq 
        -2 \lambda_{con} V + \Bar{d} \sqrt{V}
    \end{align}
    where     
    \begin{equation}
        \lambda_{con} = \frac
            {\min[k, \frac{\lambda}{2}(\lambda_\mathrm{min}(\Bar{P}^{-1}) + \gamma)]}
            {\lambda_\mathrm{max}(M)}
    \end{equation} 
    and 
    \begin{equation} 
        \Bar{d} = \sqrt{1/\lambda_\mathrm{min}(M)}
            \left\| 
            \begin{bmatrix}
                d \\ W^T d_1 + \lambda \gamma a
            \end{bmatrix}\right\| \\ 
    \end{equation}
    
    Applying the transformation $ W = \sqrt{y^T M y} $ and the comparison lemma from \cite{khalil2002nonlinearsystems}, we arrive at 
    % Consider the substitution $ W = \sqrt{y^T M y} $.  Then, 
    % \begin{align}
    %     \dot W &= \frac{1}{2 \sqrt{V}} \dot V \\
    %     &\leq \frac{-2}{2 \sqrt{V}}(\lambda_{con} V + \Bar{d} \sqrt{V}) \\
    %     &= - \lambda_{con} W + \Bar{d}
    % \end{align}
    % and we can now apply the comparison lemma to obtain
    % \begin{equation}
    %     W(t) \leq W(0) e^{-\lambda_{con} t} + \frac{\Bar{d}}{\lambda_{con}}\left(1 - e^{-\lambda_{con} t} \right)
    % \end{equation}
    % Therefore
    \begin{equation}
        \| y\| \leq \sqrt{\kappa(M)} \| y(0) \| e^{-\lambda_{con} t} + \frac{\Bar{d}}{\lambda_{con}\sqrt{\lambda_\mathrm{min}M}}\left(1 - e^{-\lambda_{con} t} \right)
    \end{equation}
    thus proving the exponential convergence to the error ball given in (\ref{equ:exponential-bound}).
\end{proof}

%% file: sections/experiments.tex
\section{{Experimental Validation}}
We implemented and tested our learning-based composite-adaptation controller on an Intel Aero Ready to Fly Drone and tested it with three different trajectories for each of three different kernel functions.  In each test, wind conditions were generated using CAST's open air wind tunnel, pictured in Fig.~\ref{fig:wind-tunnel}.  The first test had the drone hover in increasing wind speeds.  The second test had the drone move quickly between different set points with increasing wind speeds.  These time varying wind conditions showed the ability of the controller to adapt to new conditions in real time.  The the third test had the drone fly in a figure 8 pattern in constant wind to demonstrate performance over a dynamic trajectory.  The prediction error, composite velocity tracking error, and position tracking error for each kernel and each trajectory is listed in Fig.~\ref{tbl:error}.

The Intel Aero Drone incorporates a PX4 flight controller with the Intel Aero Compute Board, which runs Linux on a 2.56GHz Intel Atom x7 processor with 4 GB RAM.  The controller was implemented on the Linux board and sent thrust and attitude commands to the PX4 flight controller using MAVROS software.  CAST's Optitrack motion capture system was used for global position information, which was broadcast to the drone via wifi.  On EKF running on the PX4 controller filtered the IMU and motion capture information, to produce position and velocity estimates.  

The CAST open air wind tunnel consists of approximately 1,400 distributed fans, each individually controllable, in a 3 by 3 meter grid.

\subsection{Data collection and Kernel Training}

\input{figures/plot-data-fashift.tex}
\input{figures/fig-validate-kernels.tex}

Position, velocity, acceleration, and motor speed data was gathered by flying the drone on a random walk trajectory for at 0, 1.3, 2.5, 3.7, and 4.9 m/s wind speeds for 2 minutes each, to generate training data.  The trajectory was generated by randomly moving to different set points in a predefined cube centered in front of the wind tunnel.  Then, using the dynamics equations defined previously, we computed the aerodynamic disturbance force, $f$.

Three different kernel functions were used in the tests.  The first was an identity kernel, $\varphi \equiv I$.  Note that with a only a tracking error update term in the adaptation law, this would be equivalent to integral control.  The second and third kernels were the vector and scalar kernels, defined in (\ref{equ:meta-1-kernels}) and (\ref{equ:meta-2-kernels}), respectively. 

During offline training, the kernels were validated by estimating $a$ using a least squares estimator on a continuous segment of a validation trajectory.  Then, the predicted force was compared to the measured force for another part of the validation trajectory.  This can be seen in Fig.~\ref{fig:validate-kernel}. 

% \subsection{Visualization of learned kernels}

\subsection{Hovering in Increasing Wind}

\input{figures/plot-hover-all.tex}

In this test, the drone was set to hover at a fixed height centered in the wind tunnel test section.  The wind tunnel was set to 2.5 m/s for 15 seconds, then 4.3 m/s for second 10 seconds, then 6.2 m/s for 10 seconds.  %Note that the first speed was one of the speeds used to generate training data, the second speed was not used to generate training data but was within the range of the training data wind speeds, and the last speed was faster than any of the training data.  
The results from this test are shown in Fig.~\ref{fig:plot-hover-all}.  

In this test we see that each controller achieves similar parameter convergence.  The facts that for each case, as the drone converges to the target hover position, the kernel functions approach a constant value, and each uses the same adaptation law, probably leads to similar convergence properties for each controller.  Note however, as seen in Fig.~\ref{tbl:error}, that the prediction error is lower for the learned kernels, as the learned kernels likely capture some of the variation in aerodynamic forces with the change in state.  

\subsection{Random Walk with Increasing Wind}

The second test had the drone move quickly between random set points in a cube centered in front of the wind tunnel, with a new set point generated every second for 60 seconds.  For the first 20 seconds, the wind tunnel was set to 2.5 m/s, for the second 20 seconds, 4.3 m/s, and for the last 20 seconds, 6.2 m/s.  The random number generator seed was fixed before each test so that each controller received the exact same set points.  

Note that the desired trajectory for this test had sudden changes in desired acceleration and velocity when the set point was moved.  Thus, the composite velocity error is significantly higher than in the other tests.  The learned kernel methods in both cases outperformed the constant kernel method in prediction error performance, but all three methods had similar tracking error performance, as seen in Figure \ref{tbl:error}.  A histogram of the prediction error at each time step the control input was calculated is shown in Figure \ref{fig:plot-setpoint-prederr-hist}.  Here we can see a slight skew of the constant kernel to higher prediction error as compared to the learned kernel methods.  Not shown in the plots here, we also noticed that for each choice of kernel, at several points throughout the test the input was completely saturated due to the discontinuities in the desired trajectory. 

\input{figures/plot-setpoint-prederr-hist.tex}

\subsection{Figure 8 with Constant Wind}

\input{figures/plot-fig8-xz.tex} 

The third trajectory was a figure 8 pattern oriented up and down (z-axis) and towards and away from the wind tunnel (x-axis).  This test used a fixed wind speed of 4.3 m/s.  In each test, the drone was started from hovering near the center of the figure 8.  Then, the wind tunnel was turned on and allowed to begin ramping up for 5 seconds, and the figure 8 trajectory was flown repeated for one minute.  Each loop around the figure 8 took 8 seconds.  

In this test, we see the most striking difference between the prediction error performance of the learned kernels versus the constant kernel, as seen in Table \ref{tbl:error}.  Despite the significantly better prediction error performance however, the velocity and position tracking error were still comparable between the three kernel choices, constant kernel, vector kernel, and scalar kernel.  Examining the estimated xz position for each test, plotted in Figure \ref{fig:plot-fig8-xz}, can offer some additional insight.  For each controller, the first one to two laps around the figure 8 pattern show large tracking errors.  However, after the parameters converge to a limit cycle value, each controller is unable to turn fast enough around the corners of the figure 8.  This suggests that the tracking error in the attitude controller is causing the position tracking performance degradation.  Two possible causes of this could be that the trajectory required quick turning speed, causing input saturation, or as a result of disturbance torques from aerodynamic effects associated with the wind.  

\subsection{Results}

Three key error metrics are given for each kernel and trajectory in Fig.~\ref{tbl:error}.  The learned kernels were able to consistently outperform the constant kernel function in prediction error, suggesting that the learned kernels were able to capture part of the position and velocity dependence of the aerodynamic effects.  This trend becomes more pronounced in the more dynamic trajectories.  However, in each test, similar composite velocity and position tracking error is seen.  

\input{figures/tbl-error.tex}

%% file: figures/plot-data-fashift.tex
\begin{figure}[t]
\includegraphics[width=\linewidth]{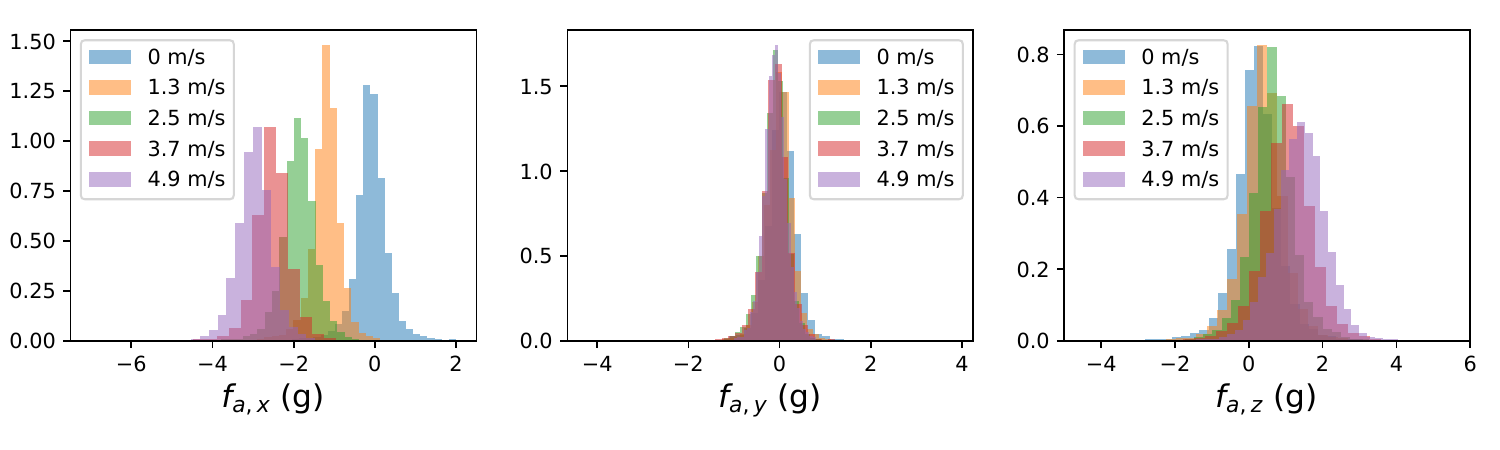}
\caption{Distributions of collected data. There are significant shifts in $f_{a,x}$ and $f_{a,z}$ under different wind speed.}
\label{fig:shift}
\end{figure}

%% file: figures/fig-validate-kernels.tex
\begin{figure}[t]
\includegraphics[width=\linewidth]{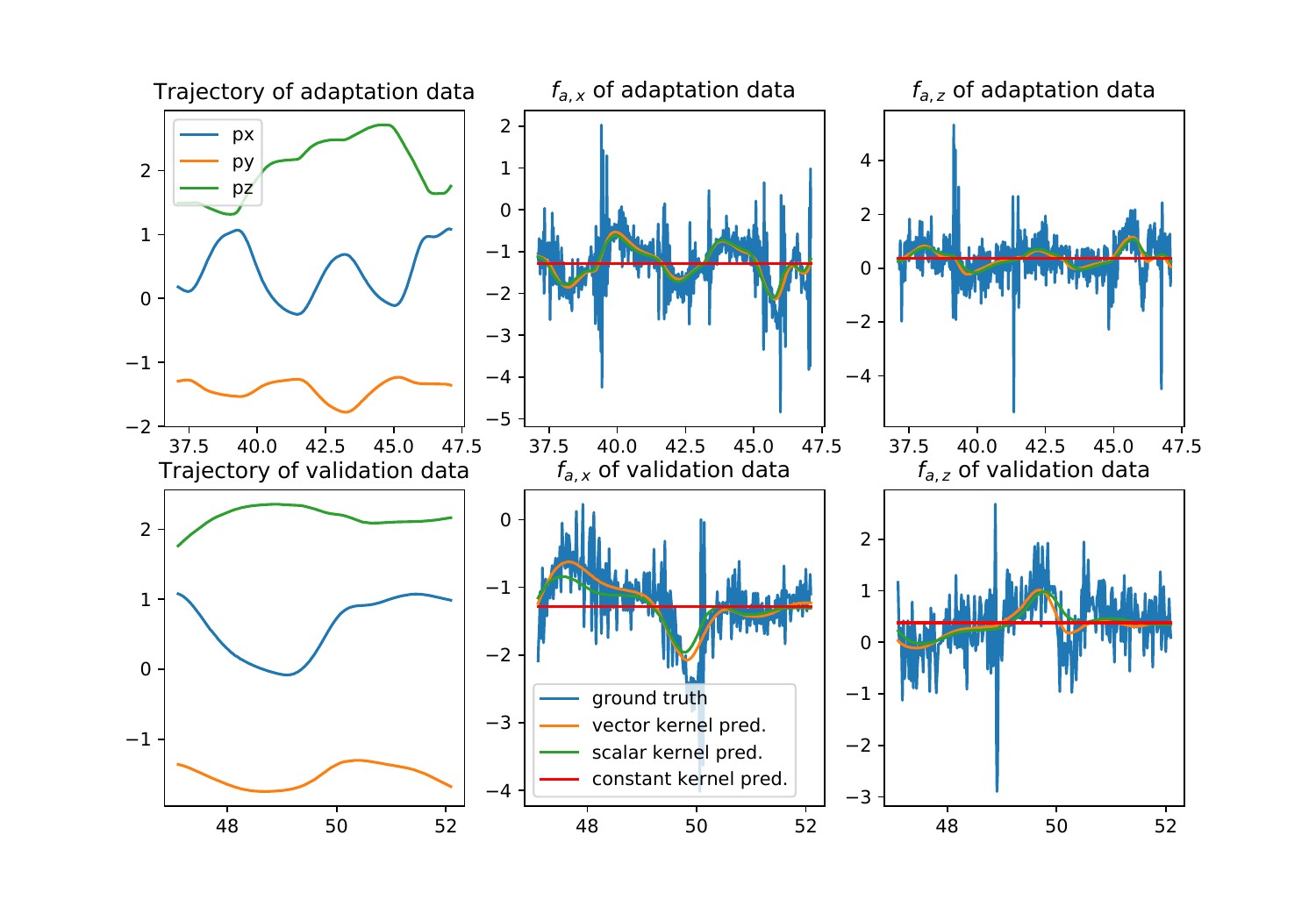}
\caption{These three plots show the validation process for each of the three kernel methods discussed in the Experiments section.  The first row shows the least squares fit of the parameters $a$ for each kernel choice.  The second row shows the prediction performance for $a$ taken from the least squares estimator in the first row.}
\label{fig:validate-kernel}
\end{figure}

%% file: figures/plot-hover-all.tex
\begin{figure}[!t]
\centering{
	{
    	\includegraphics[width=\columnwidth]{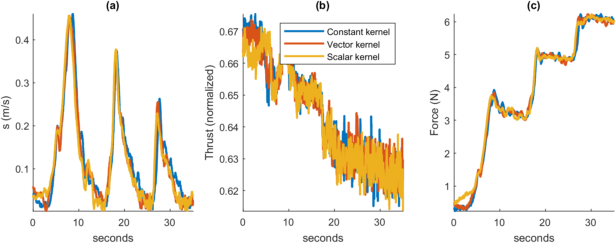}
    }
}
\caption{These plots visualize the results from the hovering in variable wind conditions.  Note the increase in wind speed can be clearly seen at 0, 15, and 25 seconds in each plot.  Plot (a) shows each choice of kernel funciton leads to similar composite velocity tracking error convergence.  Plot (b) shows that the drone gained efficiency and required a lower throttle to maintain hover as wind speed increased.  Plot (c) shows the increase in the magnitude of the measured aerodynamic force as the wind speed increased.  }
\label{fig:plot-hover-all}
\end{figure}

%% file: figures/plot-setpoint-prederr-hist.tex
\begin{figure}[!t]
\centering{
	{
    	\includegraphics[width=0.5\columnwidth]{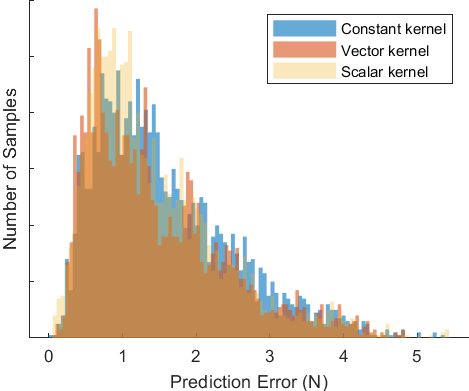}
    }
}
\caption{Histogram of prediction error for each choice of kernel in set point tracking test.  A slight bias of constant kernel towards higher prediction error can be seen here.}
\label{fig:plot-setpoint-prederr-hist}
\end{figure}

%% file: figures/plot-fig8-xz.tex
\begin{figure}[!t]
\centering{
	{
    	\includegraphics[width=0.9\columnwidth]{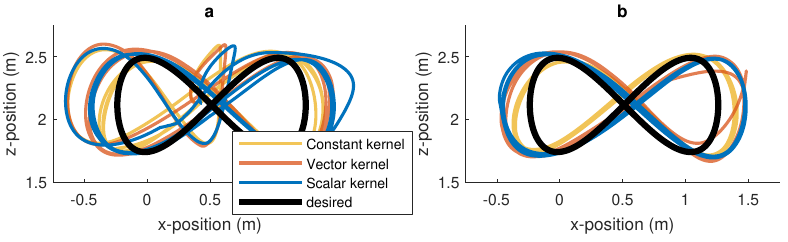}
    }
}
\caption{In plot (a), the first four laps around the figure 8 trajectory are shown.  The parameters were initialized to $0$ for each kernel, leading to very high initial tracking error.  For each choice of kernel, the tracking error converged near some limit cycle behavior, as seen in plot (b).}
\label{fig:plot-fig8-xz}
\end{figure}

%% file: figures/tbl-error.tex
\begin{figure}[!t]
\begin{center}
    \small
    \begin{tabular}{c c | c c c} 
        \multirow{2}{1cm}{Trajectory} & \multirow{2}{1cm}{Kernel} & \multicolumn{3}{|c}{Mean error metric} \\ %[0.5ex] 
        && $\|e_1\|$ (N) & $\|s\|$ (m/s) & $\|q-q_d\|$ (m) \\ 
        \hline
        \multirow{3}{1cm}{Hover}& constant  & 0.75 & 0.11 & 0.13 \\ 
        & vector & 0.72 & 0.12 & 0.13 \\
        & scalar & 0.71 & 0.11 & 0.12 \\
        \hline
        \multirow{3}{1cm}{Set point}& constant  & 1.58 & 0.63 & 0.25 \\ 
        & vector & 1.45 & 0.61 & 0.24 \\
        & scalar & 1.42 & 0.60 & 0.24 \\
        \hline
        \multirow{3}{1cm}{Figure 8} & constant  & 0.91 & 0.24 & 0.20 \\ 
        & vector & 0.74 & 0.26 & 0.22 \\
        & scalar & 0.75 & 0.25 & 0.21 \\
    \end{tabular}
\end{center}
\caption{Prediction error ($\mathrm{mean}(\|e_1\|)$), composite velocity error ($\textrm{mean}(\|s\|)$), and position tracking error ($\textrm{mean}(\|q - q_d\|)$) for each kernel function and trajectory tested}
\label{tbl:error}
\end{figure}

%% file: sections/discussion.tex
\section{{Results and Discussion}}

In this paper we have presented an integrated approach that uses prior data to develop a drone controller capable of adapting to new and changing wind conditions.  A meta-learning formulation to the offline training helped us design kernel functions that can represent the dynamics effects observed the training data.  Then we designed an adaptive controller that can exponentially stabilize the system.

In the our experiments, we saw the the learned kernels were able to reduce prediction error performance over a constant kernel.  However, this did not translate into improved tracking error performance.  We believe that this could be caused by a combination of attitude tracking error, input saturation, and dependence of unmodelled dynamics on control input.  In our tests we saw both input saturation and attitude tracking error lead to increased position tracking error.  We also know that different aerodynamic effects can cause a change in rotor thrust, usually modelled as a change in the coefficient of thrust.

All of our tests have demonstrated the viability of our approach. Our tests have lead us to two conclusions.  First, for the tests we ran, the adaptive control formulation (with either constant kernel or learned kernel) is able to effectively compensate for the unmodelled aerodynamic effects and adapt to changing wind conditions in real time.  Second, we have demonstrated our approach to incorporate learned dynamics into a robust control design.  

%Further work will include implementing this method on %attitude control, accounting for input saturation, and %learning dependence on control input.  LET's comment it out